\documentclass[11pt,a4paper]{article}
\usepackage[hyperref]{acl2021}
\usepackage{times}
\usepackage{latexsym}

\usepackage{microtype}      % make text margins prettier
\usepackage{graphicx}       % import graphical images
\usepackage[T1]{fontenc}    % fix font related glitches such as "|"
\usepackage{multirow}       % merge rows in tables
\usepackage{subcaption}     % create sub-tables and sub-figures
\usepackage{amssymb}        % enable \mathbb, \mathcal
\usepackage{bold-extra}     % enable \texttt{\textbf{}}
\usepackage{bm}             % enable bold font in math mode
\usepackage[hang,flushmargin]{footmisc}  % minimize footnote indentation
\usepackage{amsmath}
\usepackage{amsfonts}
\usepackage{caption}
\usepackage{paralist}
\newcommand{\LN}{\linebreak\noindent}    % to manage inline spacing
\newcommand{\textsec}[1]{\textsection\ref{#1}}
\DeclareMathOperator*{\argmax}{arg\,max}

\aclfinalcopy % Uncomment this line for the final submission
\title{Levi Graph AMR Parser using Heterogeneous Attention}

\author{Han He \\
  Computer Science \\
  Emory University \\
  Atlanta GA 30322, USA \\
  \texttt{han.he@emory.edu} \\\And
  Jinho D. Choi \\
  Computer Science \\
  Emory University \\
  Atlanta GA 30322, USA \\
  \texttt{jinho.choi@emory.edu} \\}

\begin{document}
\maketitle

\begin{abstract}

Coupled with biaffine decoders, transformers have been effectively adapted to text-to-graph transduction and achieved state-of-the-art performance on AMR parsing.
Many prior works, however, rely on the biaffine decoder for either or both arc and label predictions although most features used by the decoder may be learned by the transformer already.
This paper presents a novel approach to AMR parsing by combining heterogeneous data (tokens, concepts, labels) as one input to a transformer to learn attention, and use only attention matrices from the transformer to predict all elements in AMR graphs\LN (concepts, arcs, labels).
Although our models \footnote{Resources are publicly available at \url{https://github.com/emorynlp/levi-graph-amr-parser}.} use significantly fewer parameters than the previous state-of-the-art graph parser, they show similar or better accuracy on AMR 2.0 and 3.0.

% Biaffine parsers have been popular for a long time due to their high efficiency. However, they suffer from error propagation as arcs are predicted independently. In this paper, we show that attention mechanism itself can generate reasonable good AMR without using any biaffine decoder. Our work starts with simplifying the Graph{\small $\leftrightarrows$}Sequence Iterative Inference (GSII) model \cite{cai-lam-2020-amr} by merging its iterative decoders into one Graph Transformer. On top of our simplified architecture, we for the first time propose a novel Levi graph parser which can generate concepts, arcs and labels solely reliant on attention. Despite its simplicity, our approaches yield comparable accuracy to the state-of-the-art parsers on AMR 2.0 and achieve superior performance on the larger AMR 3.0 dataset. Our source codes and trained models are publicly available.\footnote{\url{https://github.com/anonymous}}
% %With the help of Electra \cite{clark2020electra}, we outperforms GSII by $x$\%.
\end{abstract}
\section{Introduction}
\label{sec:introduction}

Abstract Meaning Representation (AMR) has recently gained lots of interests due to its capability in capturing abstract concepts \cite{banarescu2013abstract}. 
In the form of directed acyclic graphs (DAGs), an AMR graph consists of nodes as concepts and edges as labeled relations.
To build such a graph from plain text, a parser needs to predict concepts and relations in concord.

While significant research efforts have been conducted to improve concept and arc predictions, label prediction has been relatively stagnated. 
Most previous models have adapted the biaffine decoder for label prediction \cite{lyu-titov-2018-amr, zhang-etal-2019-amr, cai-lam-2019-core, zhou-etal-2020-amr, lindemann-etal-2020-fast}.
These models assign labels from the biaffine decoder to arcs predicted by another decoder, which can be misled by incorrect arc predictions during decoding.

The enhancement of message passing between decoders for arc and label predictions has shown to be effective.
Among these works, \citet{cai-lam-2020-amr} emerge with an iterative method to exchange embeddings between concept and arc predictions\LN and feed the enhanced embeddings to the biaffine decoder for label prediction. 
While this approach greatly improves accuracy, it complicates the network architecture without structurally avoiding the error propagation from the arc prediction.

This paper presents an efficient transformer-based \cite{vaswani2017attention} approach that takes a mixture of tokens, concepts, and labels as inputs, and performs concept generation, arc prediction, and label prediction jointly using only attentions from the transformer without using a biaffine decoder. Its compact structure (\textsec{ssec:concept-biaffine-attention}) enables cross-attention between heterogeneous inputs, providing a complete view of the partially built graph and a better representation of the current parsing state.
A novel Levi graph decoder (\textsec{ssec:concept-label-attention}) is also proposed that reduces the number of decoder parameters by 45\% (from 5.5 million to 3.0 million) yet gives similar or better performance.
To the best of our knowledge, this is the first text-to-AMR graph parser that operates on the heterogeneous data and adapts no biaffine decoder.

\section{Related Work}
\label{sec:related-work}

Recent AMR parsing approaches can be categorized into four classes:
\begin{inparaenum}[(i)]
\item transition-based parsing which casts the parsing process into a sequence of transitions defined on an abstract machine (e.g., a transition system using a buffer and a stack) \cite{wang-etal-2016-camr, damonte-etal-2017-incremental, ballesteros-al-onaizan-2017-amr, peng-etal-2017-addressing, guo-lu-2018-better, liu-etal-2018-amr, naseem-etal-2019-rewarding, fernandez-astudillo-etal-2020-transition, lee-etal-2020-pushing},
\item seq2seq-based parsing \footnote{Seq2seq-based parsing is sometimes categorized into ``translation-based methods'' \cite{koller-etal-2019-graph} possibly due to the prevalence of seq2seq model in Neural Machine Translation, while we believe that translation refers more to the transduction between languages while AMR is neither a language nor an interlingua. } which transduces raw sentences into linearized AMR graphs in text form \cite{barzdins-gosko-2016-riga, konstas-etal-2017-neural, van2017neural, peng-etal-2018-sequence, xu-etal-2020-improving, bevilacqua-etal-2021-one},
\item seq2graph-based parsing which incrementally and directly builds a semantic graph via expanding graph nodes without resorting to any transition system \cite{cai-lam-2019-core, zhang-etal-2019-broad, lyu2020differentiable}.
\item graph algebra parsing which translates an intermediate grammar structure into AMR \cite{artzi-etal-2015-broad, groschwitz-etal-2018-amr, lindemann-etal-2019-compositional, lindemann-etal-2020-fast}.
\end{inparaenum}  

Our work is most closely related to seq2graph paradigm while we extend the definition of node to accommodate relation labels in a Levi graph. We generate a Levi graph which is a linearized form originally used in seq2seq models for AMR-to-text \cite{beck-etal-2018-graph, guo-etal-2019-densely, ribeiro-etal-2019-enhancing}. Our Levi graph approach differs from seq2seq approaches in its attention based arc prediction, where arc is directly predicted by attention heads instead of brackets in the target sequence.
\section{Approach}
\label{sec:approach}

% Our model predicts one node in the graph at a time.
%Concept is the node 
%target node is the one to be predicted
% label means R
%leading to a universal parsing architecture without the use of an biaffine label prediction layer.

%%%%%%%%%%%%%%%%%%%%%%%%%%%%%%%%%%%%%%%%%%%%%%%%%%

\subsection{Text-to-Graph Transducer}
\label{ssec:text-to-graph-transducer}

\begin{figure}[htbp!]
\centering
\includegraphics[width=\columnwidth]{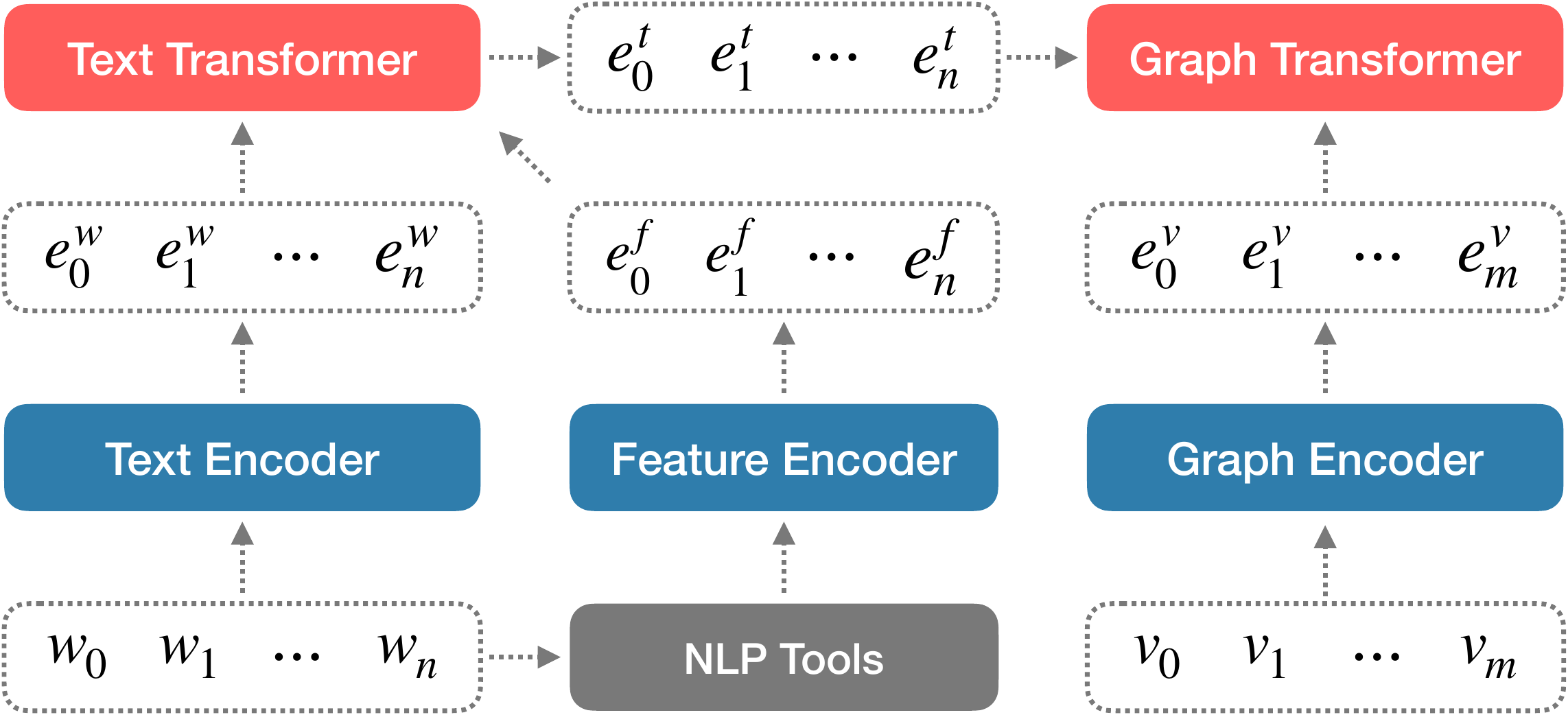}
\caption{Overview of our Text-to-Graph Transducer.}
\label{fig:text-to-graph-transducer}
\end{figure}

\noindent Figure~\ref{fig:text-to-graph-transducer} shows the overview of our Text-to-Graph Transduction model.
Let $W = \{w_0, w_1, \ldots, w_{n}\}$ be the input sequence where $w_0$ is a special token representing 
the target node and $w_i$ is the $i$'th token.
$W$ is fed into a \textit{Text Encoder} creating embeddings $\{e^w_0, e^w_1, \ldots, e^w_{n}\}$.
In parallel, \textit{NLP Tools} produce several features for $w_i$ and pass them to a \textit{Feature Encoder} to generate $\{e^f_0, e^f_1, \ldots, e^f_{n}\}$.
Embeddings $\{e^{w}_i \oplus e^f_i : i \in [0, n]\}$ are put to a \textit{Text Transformer}, which generates $E^t = \{e^t_0, e^t_1, \ldots, e^t_n\}$.\footnote{In our case, \texttt{BERT} \cite{devlin-etal-2019-bert} is used as the \textit{Text Encoder} and $\forall_i. e^f_i = e^{\textsc{lemma}}_i \oplus e^{\textsc{pos}}_i \oplus e^{\textsc{ner}}_i \oplus e^{\textsc{char}}_i$ is created by the \textit{Feature Encoder} using predictions (lemmas, part-of-speech tags and named-entities) from the \textit{NLP Tools} and character level features from a Convolutional Neural Network. In this work, we use CoreNLP \cite{manning2014stanford} for a fair comparison with existing approaches.}

\noindent Let $V = \{v_0, v_1, \ldots, v_m\}$ be the output sequence where $v_0$ is a special token representing the root and $v_i$ is the $i$'th predicted node.
$V$ is fed into a \textit{Graph Encoder} to create $E^v = \{e^v_0, e^v_1, \ldots, e^v_{m}\}$.
Finally, $E^t$ and $E^v$ are fed into a \textit{Graph Transformer} that predicts the target node as well as its relations to all nodes in $V$.
The target node predicted by the \textit{Graph Transformer} gets appended to $V$ afterwards.\footnote{\textit{Graph Encoder} creates $\forall_i. e^v_i = \text{transformer}(e^{\textsc{node}}_i \oplus e^{\textsc{char}}_i$).}

%%%%%%%%%%%%%%%%%%%%%%%%%%%%%%%%%%%%%%%%%%%%%%%%%%

\subsection{Concept + Arc-Biaffine + Rel-Biaffine}
\label{ssec:concept-biaffine}

Our first graph transformer generates $\{v_1, \ldots, v_m\}$ where $v_i$ is a concept in the target graph, and predicts both arcs and labels using a biaffine decoder.
Given $E^t$ and $E^v$ (\textsec{ssec:text-to-graph-transducer}), three matrices are created, $\mathcal{Q} = e^t_0 \in \mathbb{R}^{1 \times d}, \mathcal{K}|\mathcal{V} = [e^t_1, .., e^t_n, e^v_0, e^v_1, .., e^v_m]$ $\in \mathbb{R}^{k \times d}$ ($k = n + m + 1$).
These matrices are put to multiple layers of multi-head attention (\texttt{MHA}) producing $\{\alpha^i : i \in [1, h]\}$ and $\{\beta^i : i \in [1, h]\}$ from\LN the last layer, where $h$ is the total number of heads in \texttt{MHA} ($\text{W}^{\mathcal{Q}|\mathcal{K}|\mathcal{V}}_i \in \mathbb{R}^{d \times d}, \text{W}^\oplus \in \mathbb{R}^{(h\cdot d) \times d}$):
\begin{align*}
\alpha^i &= \text{softmax}(\frac{(\mathcal{Q}\text{W}^\mathcal{Q}_i)(\mathcal{K}\text{W}^\mathcal{K}_i)^\top}{\sqrt{d}}) &\in\mathbb{R}^{1\times k}\\
\beta^i &= \alpha^i \cdot \mathcal{V} \cdot \text{W}^\mathcal{V}_i &\in \mathbb{R}^{1\times d}\\
\alpha^\oslash &= [\alpha^1_j : j \in [1, n]] &\in \mathbb{R}^{1 \times n} \\
\beta^\oplus &= (\beta^1 \oplus \ldots \oplus \beta^h) \cdot \text{W}^\oplus &\in \mathbb{R}^{1\times d}\\
\end{align*}
$\alpha^\oslash_j$ indicates the probability of $w_j$ being aligned to the target node, and $\beta^\oplus$ is the embedding representing the node.
Let $C$ be the list of all concepts in training data and $L$ be the list of lemmas for tokens in $W$ such that $|W| = |L|$.
Given $X = C^\frown W^\frown L$, $\alpha^\oslash$ and $\beta^\oplus$ are fed into a \textit{Node Decoder} estimating the score of each  $x_i \in X$ being the target node:
\begin{align*}
g(C|W|L) &= \text{softmax}(\beta^\oplus \cdot \text{W}^{C|W|L})\\
p(x_i)   &= g(C) \cdot [\text{softmax}(\beta^\oplus \cdot \text{W}^G)]_i\\
         &+ g(W) \sum_{j \in W(x_i)} \alpha^\oslash_j + g(L) \sum_{j \in L(x_i)} \alpha^\oslash_j
\end{align*}
$g(C|W|L)$ is the gate probability of the target node being in $C|W|L$, respectively ($\text{W}^{C|W|L} \in \mathbb{R}^{d \times 1}$).
$p(x_i)$ is estimated by measuring the probabilities of $x_i$ being the target if $x_i \in C$ ($\text{W}^G \in \mathbb{R}^{d \times |C|}$), and if $x_i \in W|L$ where $W|L(x_i) = \{j : (x_i = y_j)\:\land y_j \in W|L\}$, respectively.
Finally, the output layer\LN $o^{\text{node}} = [p(x_i) : x_i \in X] \in \mathbb{R}^{1 \times (|C|+|W|+|L|)}$ gets created and $\argmax_{x_i}(o^{\text{node}})$ is taken as the target.

\begin{figure}[htbp!]
\centering
\includegraphics[width=\columnwidth]{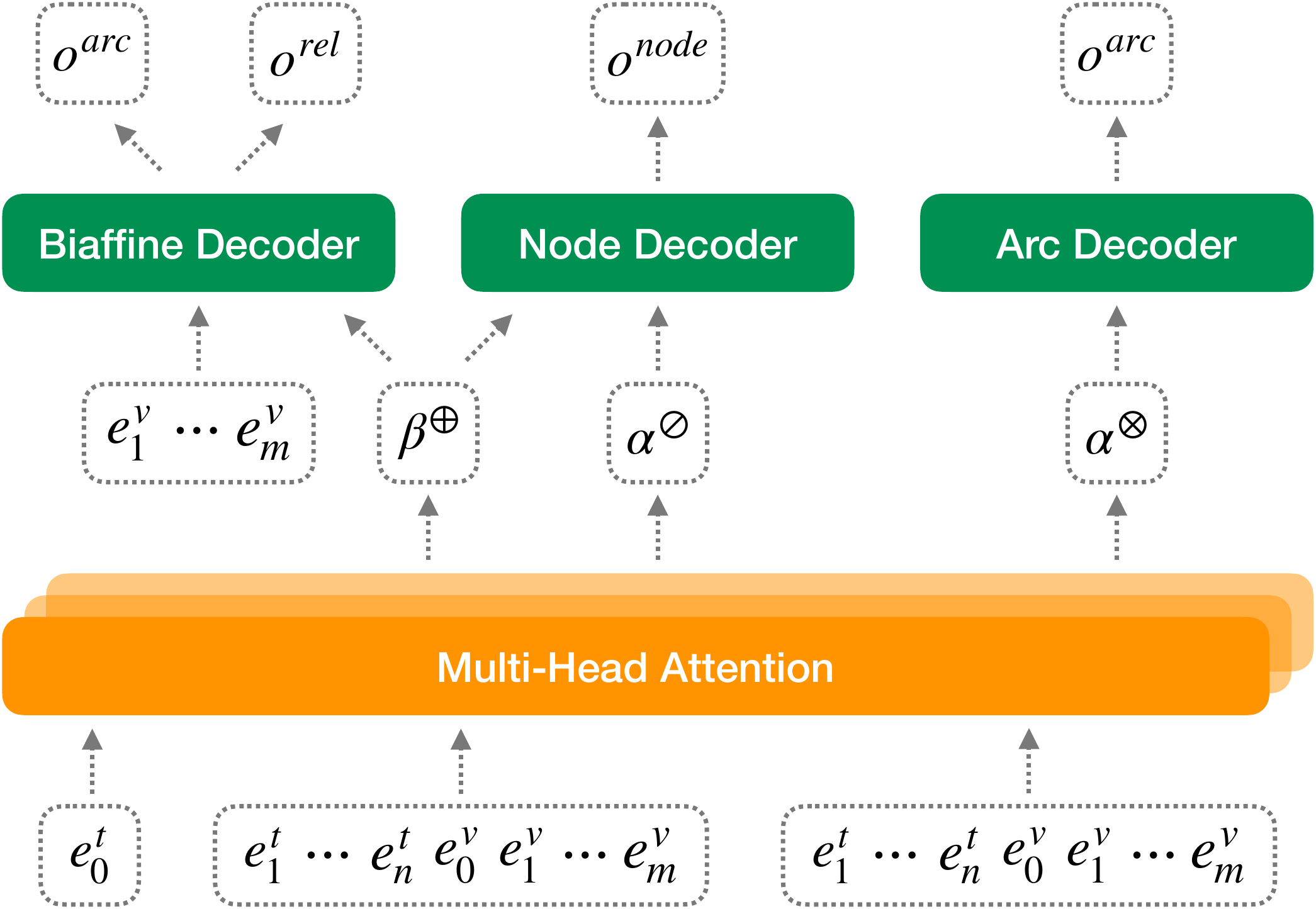}
\caption{Overview of our Graph Transformer models. \texttt{ND}/\texttt{BD}/\texttt{AD}: node/biaffine/arc decoder. \textsec{ssec:concept-biaffine}: \texttt{ND} for concept generation and \texttt{BD} for arc and label predictions; \textsec{ssec:concept-biaffine-attention}: \texttt{ND} for concept generation, \texttt{AD} for arc prediction, and \texttt{BD} for label prediction; \textsec{ssec:concept-label-attention}: \texttt{ND} for concept and label generations and \texttt{AD} for arc prediction.}
\label{fig:graph-transformer}
\end{figure}

\noindent For arc and label predictions, the target embedding $\beta^\oplus$ is used to represent a head and the embeddings of previously predicted nodes, $\{e^v_1, \ldots, e^v_m\}$, are used to represent dependents in a \textit{Biaffine Decoder}, which creates two output layers, $o^\text{arc} \in \mathbb{R}^{1 \times m} $ and $o^\text{rel} \in \mathbb{R}^{1 \times m \times |R|}$, to predict the target node being a head of the other nodes, where $|R|$ is the list of all labels in training data \cite{dozat:17a}.

%%%%%%%%%%%%%%%%%%%%%%%%%%%%%%%%%%%%%%%%%%%%%%%%%%

\subsection{Concept + Arc-Attention + Rel-Biaffine}
\label{ssec:concept-biaffine-attention}

Our second graph transformer is similar to the one in \textsec{ssec:concept-biaffine} except that it uses an \textit{Arc Decoder} instead of the \textit{Biaffine Decoder} for arc prediction.
Given $A = \{\alpha^1, \ldots, \alpha^h\}$ in \textsec{ssec:concept-biaffine}, $\alpha^\otimes \in \mathbb{R}^{1 \times (m+1)}$ is created\LN by first applying dimension-wise maxpooling to $A$ and slicing the last $m+1$ dimensions as follows:
$$
\alpha^\otimes = [\max(\alpha^1_j, \ldots, \alpha^h_j) : j \in [n+1, n+m+1]]
$$
Notice that values in $\alpha^\otimes$ are derived from multiple heads; thus, they are not normalized.
Each head is expected to learn different types of arcs.
During decoding, any $v_i \in V$ whose $\alpha^\otimes_i \geq 0.5$ is predicted to be a dependent of the target node.
During training, the negative log-likelihood of $\alpha^\otimes$ is optimized.\footnote{This model still uses the \textit{Biaffine Decoder} for label prediction.}

The target node, say $v_t$, may need to be predicted as a dependent of $v_i$, in which case, the dependency is reversed (so $v_t$ becomes the head of $v_i$), and the label is concatenated with the special tag \texttt{\_R} (e.g., $\texttt{ARG0}(v_i, v_t)$ becomes $\texttt{ARG0\_R}(v_t, v_i)$).

%%%%%%%%%%%%%%%%%%%%%%%%%%%%%%%%%%%%%%%%%%%%%%%%%%

\subsection{Levi Graph + Arc-Attention}
\label{ssec:concept-label-attention}

\begin{figure}[htbp!]
 \centering
 \begin{subfigure}[b]{0.44\columnwidth}
     \centering
     \includegraphics[width=\textwidth]{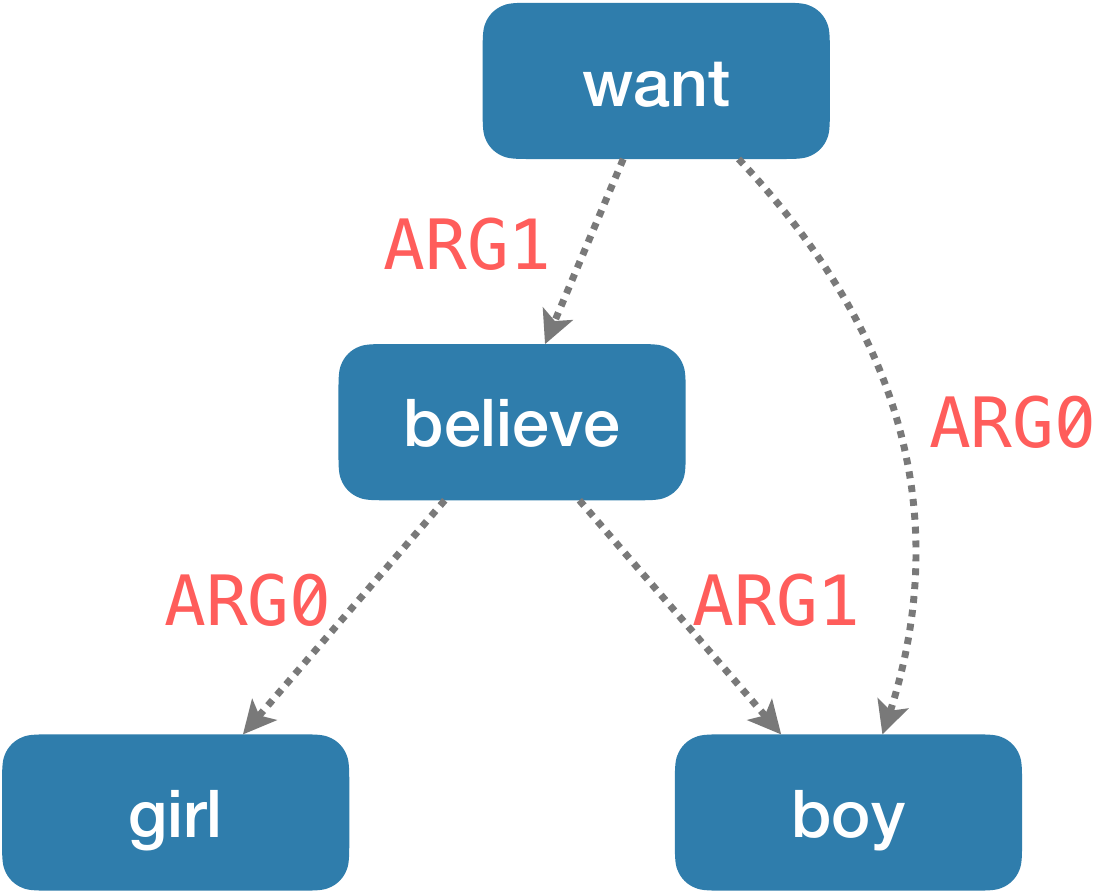}
     \caption{AMR graph}
     \label{fig:graph-amr}
 \end{subfigure}
 \hfill
 \begin{subfigure}[b]{0.50\columnwidth}
     \centering
     \includegraphics[width=\textwidth]{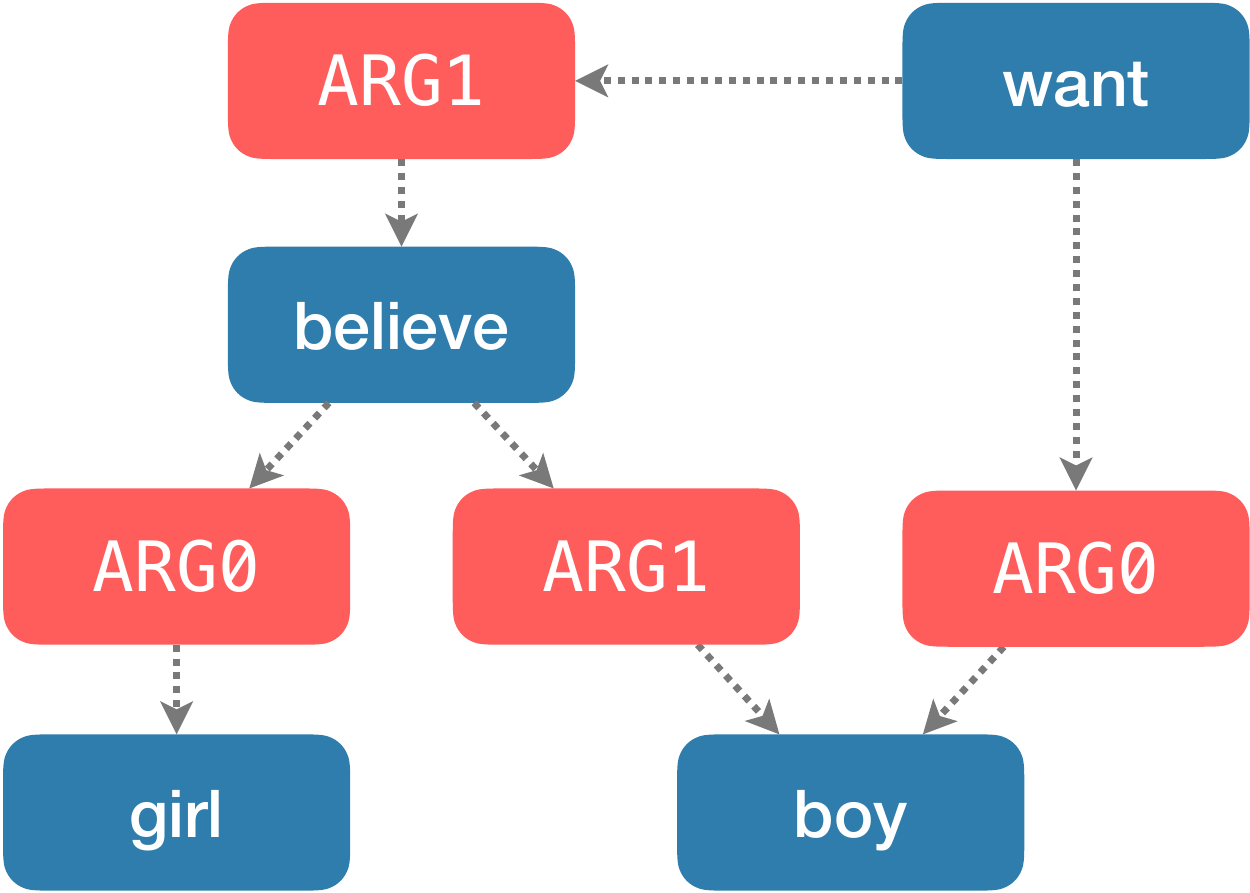}
     \caption{Levi graph}
     \label{fig:graph-levi}
 \end{subfigure}
\caption{AMR and Levi graphs for the input, ``\textit{The boy wants the girl to believe him}''.}
\label{fig:graph-amr-levi}
\end{figure}

Our last graph transformer uses the \textit{Node Decoder} for both concept and label generations and the \textit{Arc Decoder} for arc prediction.
In this model, $v_i \in V'$ can be either a concept or a label such that the original AMR graph is transformed into the Levi graph\LN \cite{levi1942finite, beck-etal-2018-graph} (Figure~\ref{fig:graph-amr-levi}).

\begin{table*}[t]
    \begin{subtable}[t]{\textwidth}
        \small
        \centering
        \resizebox{1.0\columnwidth}{!}{
			\begin{tabular}{l||l|l||l|l|l|l|l|l|l}
				& \multicolumn{2}{c||}{\textbf{\textsc{Smatch}}} & \multicolumn{7}{c}{\bf Fine-grained Evaluation}\\
				\cline{2-10}
				& \multicolumn{1}{c|}{\scriptsize\bf Labeled} & \multicolumn{1}{c||}{\scriptsize\bf Unlabeled} & \multicolumn{1}{c|}{\scriptsize\bf No WSD} & \multicolumn{1}{c|}{\scriptsize\bf Concept} & \multicolumn{1}{c|}{\scriptsize\bf SRL} & \multicolumn{1}{c|}{\scriptsize\bf Reent.} & \multicolumn{1}{c|}{\scriptsize\bf Neg.} & \multicolumn{1}{c|}{\scriptsize\bf NER} & \multicolumn{1}{c}{\scriptsize\bf Wiki}\\
				\hline\hline
%				\newcite{van2017neural} & 71.0 & 74 & 72 & 82 & 66 & 52 & 62 & 79 & 65\\ 
%				\newcite{groschwitz-etal-2018-amr} & 71.0&74&72 & 84 & 64 & 49 & 57&78&71\\
%				\newcite{lyu-titov-2018-amr} & 74.4 &77.1 & 75.5 & 85.9 & 69.8 & 52.3 & 58.4 & 86.0 & 75.7\\
				% \newcite{cai-lam-2019-core} & 73.2 & 77.0& 74.2&84.4&66.7&55.3&62.9&82.0&73.2\\
				\newcite{lindemann-etal-2019-compositional} &75.3 &-&-&-&-&-&-&-&-\\
				\newcite{naseem-etal-2019-rewarding} & 75.5 & 80 & 76 & 86 & 72 & 56 & 67 & 83& 80\\
				\newcite{zhang-etal-2019-amr} & 76.3&79.0&76.8&84.8&69.7&60.0&75.2&77.9&85.8\\
				\newcite{zhang-etal-2019-broad} & 77.0 & 80& 78& 86&71&61&77&79&86\\
				\newcite{cai-lam-2020-amr} & 80.2 & 82.8 & 80.8 & 88.1 & 74.2 & 64.6 & 78.9 & 81.1 & 86.3 \\
				\newcite{xu-etal-2020-improving}\rlap{$^\dagger$}  & 80.2 & 83.7 & 80.8 & 87.4 & 78.9 & 66.5 & 71.5 & 85.4 & 75.1 \\
				\newcite{lee-etal-2020-pushing}\rlap{$^\ddagger$}  & 81.3 & 85.3 & 81.8 & 88.7 & 88.7 & 66.3 & 79.2 & 71.9 & 79.4 \\
				\newcite{bevilacqua-etal-2021-one}\rlap{$^\S$} & 84.5 & 86.7 & 84.9 & 89.6 & 79.7 & 72.3 & 79.9 & 83.7 & 87.3 \\
				\hline
                \texttt{CL20}                             & \textbf{80.0}$\pm$0.2 & 82.5$\pm$0.3          & \textbf{80.5}$\pm$0.3 & 88.0$\pm$0.1          & 73.7$\pm$0.4          & 63.8$\pm$0.7          & 79.2$\pm$0.3          & \textbf{81.1}$\pm$0.3 & \textbf{86.2}$\pm$0.1 \\
				\texttt{ND} + \texttt{BD} + \texttt{BD}                 & 79.4$\pm$0.1          & 82.3$\pm$0.1          & 80.0$\pm$0.2          & 87.9$\pm$0.2          & 73.1$\pm$0.2          & 62.5$\pm$0.2          & \textbf{79.8}$\pm$0.3 & 80.7$\pm$1.0          & 85.8$\pm$0.5 \\
				\texttt{ND} + \texttt{AD} + \texttt{BD}   & \textbf{80.0}$\pm$0.1 & \textbf{82.6}$\pm$0.1 & \textbf{80.5}$\pm$0.1 & \textbf{88.2}$\pm$0.1 & 73.6$\pm$0.4          & 63.3$\pm$0.4          & 79.4$\pm$1.0          & 80.8$\pm$0.8          & \textbf{86.2}$\pm$0.3 \\
				\texttt{ND} + \texttt{AD} + \texttt{LV} & \textbf{80.0}$\pm$0.1 & 82.2$\pm$0.2          & \textbf{80.5}$\pm$0.1 & 87.7$\pm$0.2          & \textbf{74.5}$\pm$0.2 & \textbf{64.1}$\pm$0.3 & 78.4$\pm$1.0          & 80.5$\pm$0.8          & \textbf{86.2}$\pm$0.3 \\
		\end{tabular}}
		\caption{Results on AMR 2.0 results. Supervised{$^\dagger$}/unsupervised{$^\S$} pre-training and self-learning{$^\ddagger$}\, are orthogonal to our work.}
		\label{tbl:amr2}
    \end{subtable}
    \vspace{0.5em}
    
    \begin{subtable}[t]{\textwidth}
        \small
        \centering
        \resizebox{1.0\columnwidth}{!}{
			\begin{tabular}{l||l|l||l|l|l|l|l|l|l}
				& \multicolumn{2}{c||}{\textbf{\textsc{Smatch}}} & \multicolumn{7}{c}{\bf Fine-grained Evaluation}\\
				\cline{2-10}
				& \multicolumn{1}{c|}{\scriptsize\bf Labeled} & \multicolumn{1}{c||}{\scriptsize\bf Unlabeled} & \multicolumn{1}{c|}{\scriptsize\bf No WSD} & \multicolumn{1}{c|}{\scriptsize\bf Concept} & \multicolumn{1}{c|}{\scriptsize\bf SRL} & \multicolumn{1}{c|}{\scriptsize\bf Reent.} & \multicolumn{1}{c|}{\scriptsize\bf Neg.} & \multicolumn{1}{c|}{\scriptsize\bf NER} & \multicolumn{1}{c}{\scriptsize\bf Wiki}\\
				\hline\hline
				\newcite{lyu2020differentiable} & 75.8  & - & - & \textbf{88.0} & 72.6 & - & - & - & - \\
				\hline
                \texttt{CL20} & 76.8$\pm$0.2 & 79.9$\pm$0.2 & 77.3$\pm$0.2 & 86.3$\pm$0.2 & 73.2$\pm$0.2 & 63.4$\pm$0.2 & 72.3$\pm$1.4 & 73.0$\pm$0.5 & 79.5$\pm$0.2 \\
                \texttt{ND} + \texttt{BD} + \texttt{BD} & 75.8$\pm$0.2 & 79.0$\pm$0.1 & 76.2$\pm$0.1 & 84.6$\pm$0.2 & 72.1$\pm$0.3 & 61.7$\pm$0.4 & 72.6$\pm$0.7 & 71.6$\pm$0.3 & 78.7$\pm$0.2 \\
                \texttt{ND} + \texttt{AD} + \texttt{BD} & 76.8$\pm$0.1 & \textbf{80.1$\pm$0.1} & 77.3$\pm$0.1 & 86.5$\pm$0.2 & 73.1$\pm$0.2 & \textbf{63.6$\pm$0.2} & \textbf{73.2$\pm$0.9} & 73.0$\pm$0.2 & \textbf{79.6$\pm$0.1 }\\
                \texttt{ND} + \texttt{AD} + \texttt{LV} & \textbf{77.0$\pm$0.2} & 79.8$\pm$0.2 & \textbf{77.5$\pm$0.2} & 86.1$\pm$0.1 & \textbf{73.6$\pm$0.3} & 62.6$\pm$0.6 & 71.3$\pm$0.4 & \textbf{73.3$\pm$0.7} & 79.5$\pm$0.3 \\
		\end{tabular}}
		\caption{Results on AMR 3.0.}
		\label{tbl:amr3}
    \end{subtable}
      \caption{Averages $\pm$ standard deviations on AMR 2.0 and 3.0 . {\texttt{CL20}: results by running the original implementation of \citet{cai-lam-2020-amr} 3 times, \texttt{ND+BD+BD}: \textsec{ssec:concept-biaffine}, \texttt{ND+AD+BD}: \textsec{ssec:concept-biaffine-attention}, \texttt{ND+AD+LV}: \textsec{ssec:concept-label-attention}.}}
     \label{tbl:results}
\vspace{-1.5ex}
\end{table*}

%\noindent Unlike the node sequence containing only concepts in the AMR graph ordered by breadth-first traverse, used as the output sequence for the models in \textsec{ssec:concept-biaffine} and \textsec{ssec:concept-biaffine-attention}, the node sequence in this model is derived by applying Kahn's topological sorting algorithm \cite{kahn:62a} to the Levi graph during training.
Unlike the node sequence containing only concepts in the AMR graph ordered by breadth-first traverse, used as the output sequence for the models in \textsec{ssec:concept-biaffine} and \textsec{ssec:concept-biaffine-attention}, the node sequence in this model is derived by inserting the label of each edge after head concept during training. This concepts-labels alternation has two advantages over a strict topological order:
\begin{inparaenum}[(i)]
\item it can handle erroneous cyclic graphs,
\item it is easier to restore relations as each label is connected to its closest concept.
\end{inparaenum}
\noindent
The heterogeneous nature of node sequences from Levi graphs allows our Graph Transformer to learn attentions among 3 types of input, tokens, concepts, and labels, leading to more informed predictions.

Let $V'$ be the output sequence consisting of both predicted concepts and labels.
Let $C'$ be the set of all concepts and labels in training data.
Compared to $V$ and $C$ in \textsec{ssec:concept-biaffine}, $V'$ is about twice larger than $V$ because every concept has one or more associated labels that indicate relations to its heads.
However,\LN $C'$ is not so much larger than $C$ because the addition from the labels is insignificant to the number of concepts that are already in $C$. By replacing $V|C$ with $V'|C'$ respectively, the \textit{Node Decoder} in \textsec{ssec:concept-biaffine} can generate both concepts and labels.
$\alpha^\otimes$ in \textsec{ssec:concept-biaffine-attention} then gives attention scores among concepts and labels that can be used by the \textit{Arc Decoder} to find arcs among them.
% These prevent the \textit{Arc Decoder} from making concept-to-concept or label-to-label arc predictions, that would not form a Levi graph.
% These don't prevent such predictions, which are prevented by zeroing out concept-to-concept and label-to-label scores

\section{Experiments}
\label{sec:experiments}

\subsection{Experimental Setup}

All models are experimented on both the AMR 2.0 (LDC2017T10) and 3.0 datasets (LDC2020T02).
AMR 2.0 has been well-explored by recent work,\LN while AMR 3.0 is the latest release about $1.5$ times larger than 2.0 that has not yet been explored much.
The detailed data statistics are shown in Table~\ref{app:environmental-setup}.\ref{tab:data-stats}.
The training, development, and test sets provided in the datasets are used, and performance is evaluated with the \textsc{Smatch} (F1) \cite{cai2013smatch} as well as fine-grained metrics \cite{damonte-etal-2017-incremental}.
The same pre- and post-processing suggested by\LN \newcite{cai-lam-2020-amr} are adapted.
Section~\ref{app:hyper-parameter} gives the hyper-parameter configuration of our models.

%\paragraph{Datasets} We evaluate our models on two AMR releases: AMR 2.0 (LDC2017T10) and AMR 3.0 (LDC2020T02). AMR 2.0 is the most extensively used AMR sembank in recent work, while AMR 3.0 is the latest release with approximately 50\% more data. 
%For both datasets, we use the data split provided with them. 
%Performance is evaluated using the \textsc{Smatch} (F1) metric \cite{cai2013smatch} and the fine-grained metrics \cite{damonte-etal-2017-incremental}. 

%\paragraph{Pre- and Post-processing} Although adapting better NLP tools could potentially improve performance as suggested by \newcite{cai-lam-2020-amr}, for the sake of fair comparison, we follow the conventional pre- and post-processing steps \cite{zhang-etal-2019-amr,cai-lam-2020-amr}.
%	\paragraph{Implementation} We use $h=4$ layers of Transformer, each has $8$ attention heads. Other hyper-parameters can be found in Table \ref{tbl:hyper-param} from the Appendix.

\subsection{Results}

% num of trainable parameters
% theirs: 				25141963
% crossattn 1 biaffine:	23039179
% levi:					22148171

All our models are run three times and their averages and standard deviations are reported in Table~\ref{tbl:results}.
Compared to \texttt{CL20} using 2 transformers to decode arcs \& concepts then apply attention across them,\LN our models use 1 transformer for the \textit{Node Decoder} achieving both objectives simultaneously.
All models except for \texttt{ND+BD} reaches the same \textsc{Smatch} score of 80\% on AMR 2.0.
\texttt{ND+AD+LV} shows a slight improvement over the others on AMR 3.0, indicating that it has a greater potential to be robust with a larger dataset.
Considering that this model uses about 3M fewer parameters than \texttt{CL20}, these results are promising.
\texttt{ND+BD+BD} consistently shows the lowest scores,\LN implying the significance of modeling concept generation and arc prediction coherently for structure learning.
\texttt{ND+AD+LV} shows higher scores for \texttt{SRL} and \texttt{Reent} whereas the other models show advantage on \texttt{Concept} and \texttt{NER} on AMR 2.0, although the trend is not as noticeable on AMR 3.0, implying that the Levi graph helps parsing relations but not necessarily tagging concepts.

%The left block of Table \ref{tbl:results} shows the \textsc{Smatch} scores of our models against the previous best competitors and state-of-the-art work. Our \texttt{ND} + \texttt{BD} model performs worse for almost every metric, verifying the importance of modeling interactions between concept generation and arc prediction. Our \texttt{ND} + \texttt{AD}+ \texttt{BD} model performs comparably well with the SOTA GSII competitor, while its architecture is much simpler. As both a further simplification and a novel approach, our Levi graph model demonstrates its ability to build AMR graph of similar or slightly better quality without any biaffine layer.
	
%The right block of Table \ref{tbl:results} shows fine-grained metrics of each subtask. Our \texttt{ND} + \texttt{AD}+ \texttt{BD} model obtains the highest scores on unlabeled graph and concept generation. Our Levi graph model shows advantage for semantic role labeling but performs worse for concept generation.

\begin{figure}[htbp!]
 \centering
 \begin{subfigure}[b]{0.3\columnwidth}
     \centering
     \includegraphics[width=\textwidth]{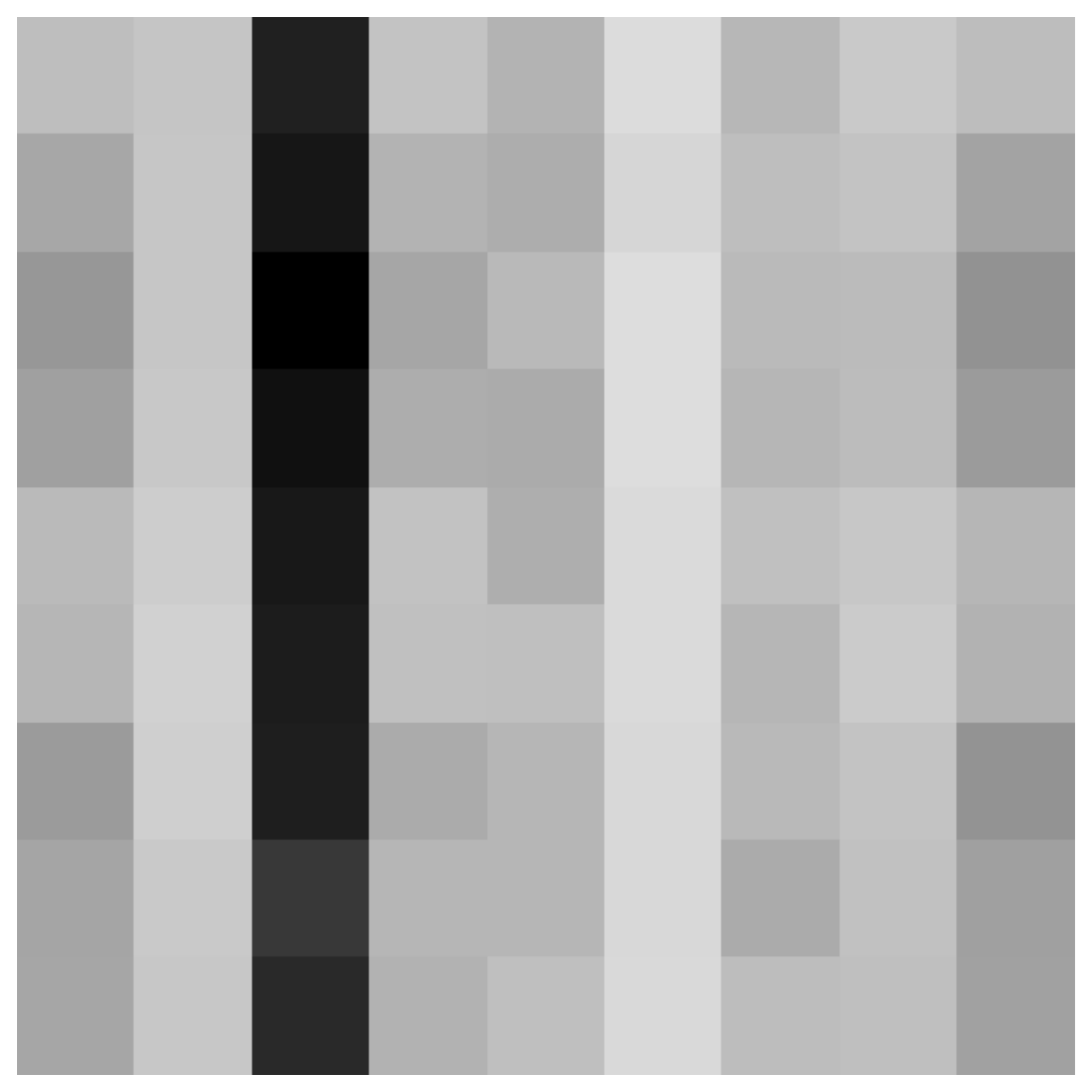}
     \caption{Token{\small $\leftrightarrows$}token }
     \label{fig:attn-token-token}
 \end{subfigure}
 \hfill
 \begin{subfigure}[b]{0.3\columnwidth}
     \centering
     \includegraphics[width=\textwidth]{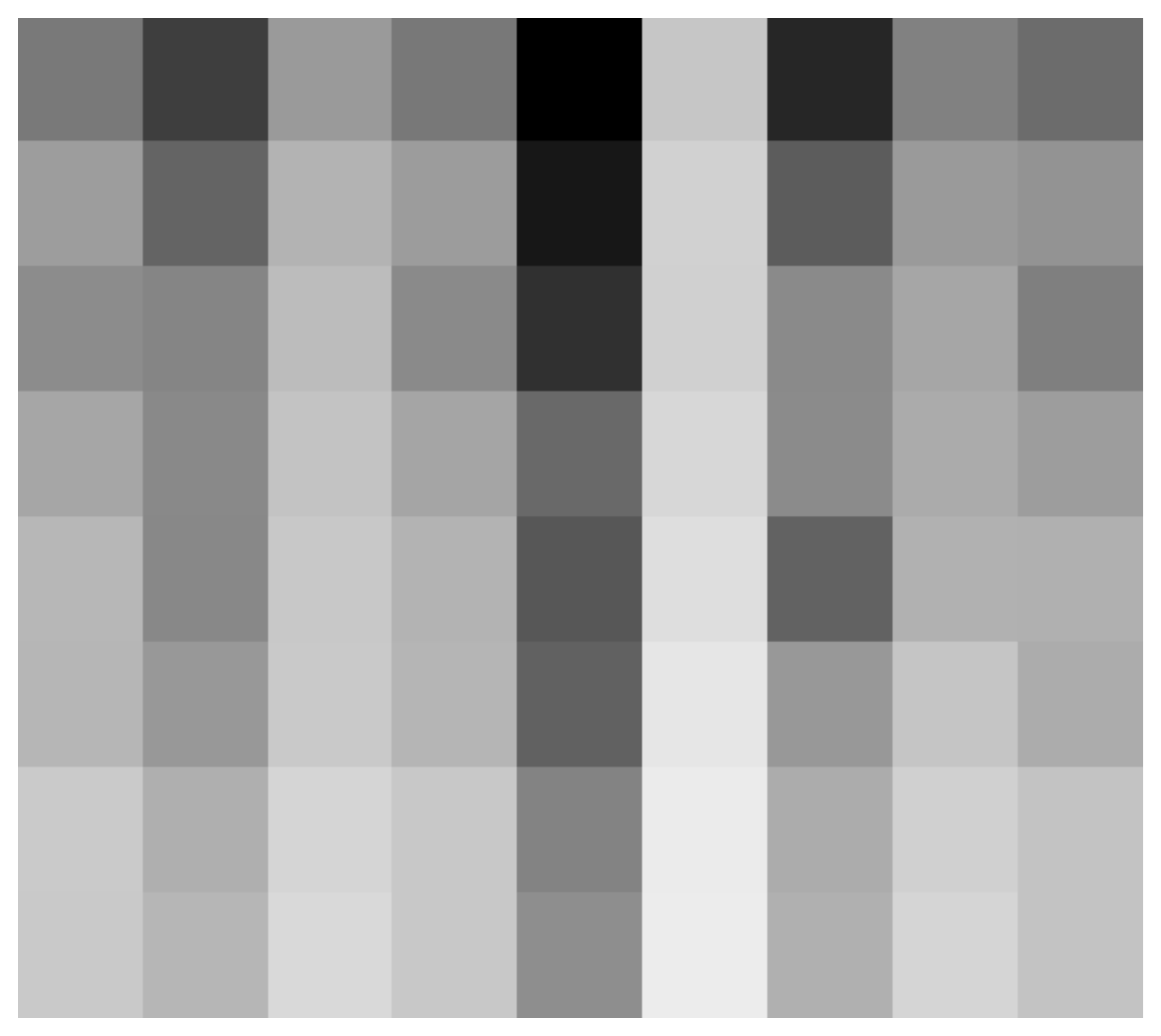}
     \caption{Node{\small $\leftrightarrows$}token }
     \label{fig:attn-node-token}
 \end{subfigure}
 \hfill
 \begin{subfigure}[b]{0.3\columnwidth}
     \centering
     \includegraphics[width=\textwidth]{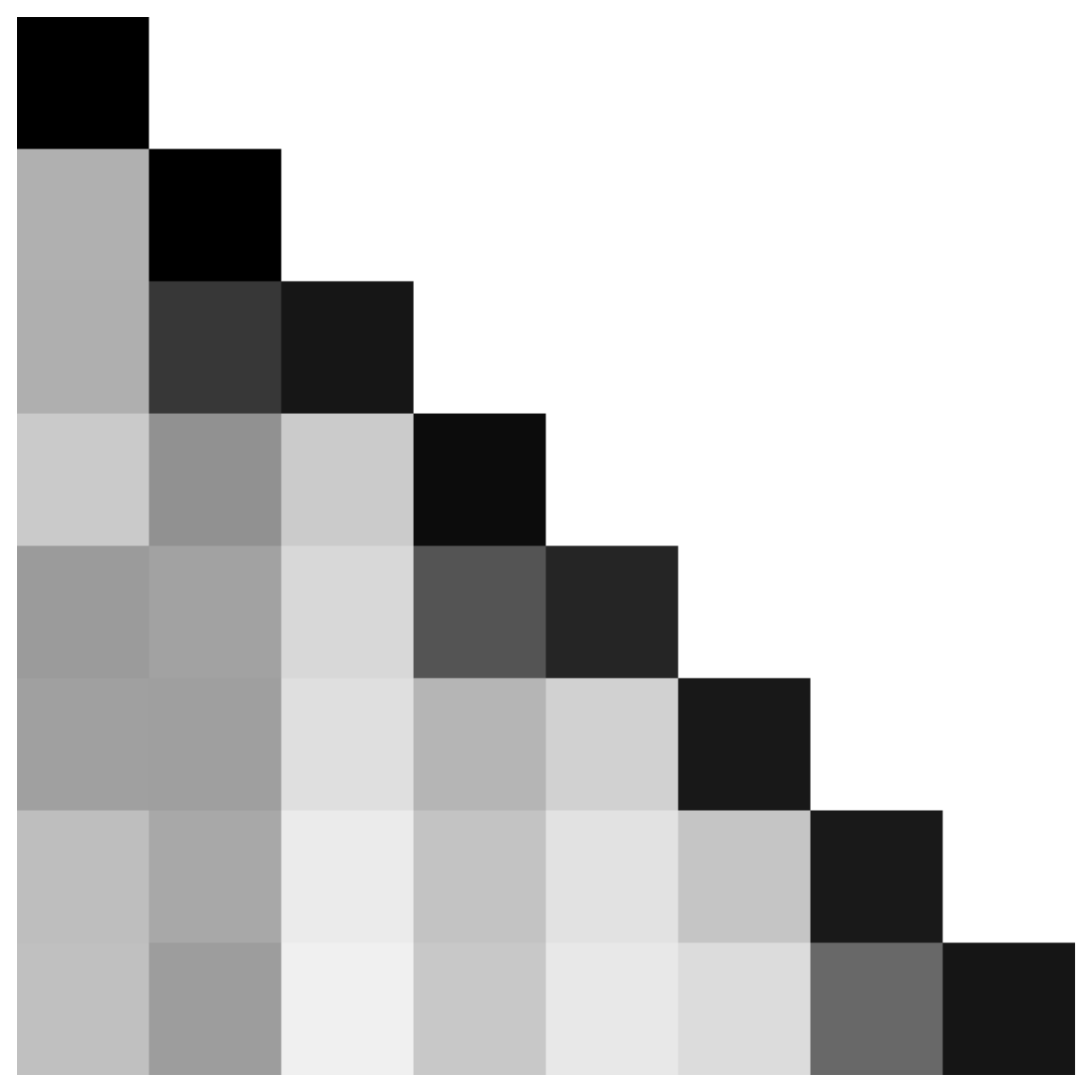}
     \caption{Node{\small $\leftrightarrows$}node}
     \label{fig:attn-node-node}
 \end{subfigure}
\caption{Self- and cross-attention for tokens ``\textit{The boy wants the girl to believe him}'' and nodes ``\textit{want believe} \texttt{ARG1} \textit{boy} \texttt{ARG1} \texttt{ARG0} \textit{girl} \texttt{ARG0}''.}
\label{fig:attn}
\vspace{-1.5ex}
\end{figure}

\paragraph{Case Study} We study the effect of our proposed two improvements: heterogeneous Graph Transformer and Levi graph, from the view of attention in Figure ~\ref{fig:attn}. Figure \ref{fig:attn-token-token} shows that the core verb ``\textit{wants}'' is heavily attended by every token, suggesting that our Graph Transformer successfully grasps the core idea. Figure \ref{fig:attn-node-token} presents the soft alignment between nodes and tokens, which surprisingly over-weights ``\textit{ boy}'', ``\textit{girl}'' and ``\textit{believe}'' possibly due to their dominance of semantics. Figure \ref{fig:attn-node-node} illustrates the arc prediction, which is a lower triangular matrix obtained by zeroing out the upper triangle of stacked $\alpha^\otimes$. Its diagonal suggests that self-loop is crucial for representing each node.

% Perhaps the following paragraph has to be removed due to page limit.
% \paragraph{Effect of Beam Size} We compare {Smatch} scores of best models with different beam sizes in Figure \ref{fig:beam}. While the best performance is always reached with beam size $8$, our approach is more accurate with greedy decoding in comparison with GSII, which suggests its potential in time-constrained scenarios. 

%\input{tex/analysis}
\section{Conclusion}
\label{sec:conclusion}

We presented two  effective approaches which achieve comparable (or better) performance comparing with the state-of-the-art parsers with significantly fewer parameters. Our text-to-graph transducer enables self- and cross-attention in one transformer, improving both concept and arc prediction. With a novel Levi graph formalism, our parser demostrates its advantage on relation labeling. An interesting future work is to preserve benefits from both approaches in one model. It is also noteworthy that our Levi graph parser can be applied to a broad range of labeled graph parsing tasks including dependency trees and many others.

\bibliography{acl2021}
\bibliographystyle{acl_natbib}

\cleardoublepage\appendix
\section{Appendix}
\label{sec:appendix}

\subsection{Datasets and Pre/Post-Processing}
\label{app:environmental-setup}

Table~\ref{tab:data-stats} describes statistics of the AMR 2.0\footnote{AMR 2.0: \url{https://catalog.ldc.upenn.edu/LDC2017T10}} and the AMR 3.0\footnote{AMR 3.0: \url{https://catalog.ldc.upenn.edu/LDC2020T02}} datasets used in our experiments.

\begin{table}[h]
    \begin{subtable}[h]{0.45\textwidth}
    	\small
        \centering
\begin{tabular}{c||c|c|c|c} 
  & \bf Sentences & \bf Tokens & \bf Concepts & \bf Relations \\
\hline\hline
\tt TRN & 36,521 & 624,750 & 422,655 & 426,712 \\
\tt DEV & 1,368 & 27,713 & 19,890 & 20,111 \\
\tt TST & 1,371 & 28,279 & 26,513 & 27,175 \\
\end{tabular}
       \caption{AMR 2.0.}
       \label{tab:amr2_stat}
    \end{subtable}
    \hfill
    \begin{subtable}[h]{0.45\textwidth}
    	\small
        \centering
\begin{tabular}{c||c|c|c|c} 
  & \bf Sentences & \bf Tokens & \bf Concepts & \bf Relations \\
\hline\hline
\tt TRN & 55,635 & 965,468 & 656,123 & 667,577 \\
\tt DEV & 1,722 & 34,696 & 25,171 & 25,568 \\
\tt TST & 1,898 & 37,225 & 34,903 & 35,572 \\
\end{tabular}
        \caption{AMR 3.0.}
        \label{tab:amr3_stat}
     \end{subtable}
\caption{Statistics of AMR 2.0 and 3.0. \texttt{TRN}/\texttt{DEV}/\texttt{TST}: training/development/evaluation set.}
\label{tab:data-stats}
\end{table}

\noindent Tokenization, lemmatization, part-of-speech and named entity annotations are generated by the Stanford CoreNLP tool \cite{manning2014stanford}.
Most\LN frequent word senses are removed and restored during pre- and post-processing.
The same graph re-categorization is performed to assign specific subgraphs to a single node as in \citet{cai-lam-2020-amr}. Wikification is done using the DBpedia Spotlight \cite{isem2013daiber} during post-processing.

\subsection{Hyper-Parameter Configuration}
\label{app:hyper-parameter}

The hyper-parameters used in our models are described in Table \ref{tbl:hyper-param}. 

\begin{table}[htbp!]
	\centering\small
	\begin{tabular}{lr}
		\hline
		\multicolumn{2}{l}{\textbf{Embeddings}} \\
		lemma & 300 \\
		POS tag& 32 \\
		NER tag & 16 \\
		concept & 300 \\
		char & 32 \\
		\hline
		\multicolumn{2}{l}{\textbf{Char-level CNN}} \\
		\#filters & 256 \\
		ngram filter size & [3] \\
		output size & 128 \\
		\hline
		\multicolumn{2}{l}{\textbf{Text Encoder}} \\
		\#transformer layers & 4 \\
		\hline
		\multicolumn{2}{l}{\textbf{Graph Encoder}} \\
		\#transformer layers & 2 \\
		\hline
		\multicolumn{2}{l}{\textbf{Transformer Layer}} \\
		\#heads & 8 \\
		hidden size & 512 \\
		feed-forward hidden size & 1024 \\
		\hline
		\multicolumn{2}{l}{\textbf{Graph Transformer}} \\
		feed-forward hidden size & 1024 \\
		\hline
		\multicolumn{2}{l}{\textbf{Biaffine}} \\
		hidden size & 100 \\
		\hline
	\end{tabular}
	\caption{Hyper-parameters settings. }
	\label{tbl:hyper-param}
\end{table}

% \subsection{Beam Search Experiments}

% \begin{figure}[htbp!]
% \centering
% \includegraphics[width=\columnwidth]{fig/beam.pdf}
% \caption{\textsc{Smatch} scores as a function of beam sizes.}
% \label{fig:beam}
% \end{figure}

% Figure ~\ref{fig:beam} shows the comparison between the best models of ours and GSII with different beam sizes.

\end{document}